\documentclass{article}

\usepackage[final,nonatbib]{neurips_2019}

\usepackage[utf8]{inputenc} 
\usepackage[T1]{fontenc}    
\usepackage{hyperref}       
\usepackage{url}            
\usepackage{booktabs}       
\usepackage{amsfonts}       
\usepackage{nicefrac}       
\usepackage{microtype}      
\usepackage{blindtext}
\usepackage{verbatim}
\usepackage{xcolor,colortbl}
\usepackage{graphicx}       
\usepackage{subcaption}
\usepackage{todonotes}

\title{Adversarial recovery of agent rewards from latent spaces of the limit order book}

\author{%
  Jacobo Roa-Vicens\\
  JP Morgan Chase \& Co.\thanks{Opinions expressed in this paper are those of the authors, and do not necessarily reflect the view of JP Morgan.}\\
  \texttt{jacobo.roavicens@jpmorgan.com}\\
  \And
  Yuanbo Wang\\
  Twitter, Inc. \thanks{Research completed at JP Morgan Chase \& Co.}\\
  \texttt{yuanbow@twitter.com}\\
  \And
  Virgile Mison\\
  JP Morgan Chase \& Co.\\
  \texttt{virgile.mison@jpmchase.com}\\
  \And
  Yarin Gal\\
  Department of Computer Science\\
  University of Oxford, United Kingdom\\
  \texttt{yarin@cs.ox.ac.uk}\\
  \And
  Ricardo Silva\\
  Department of Statistical Science\\
  University College London, United Kingdom\\
  \texttt{ricardo.silva@ucl.ac.uk}\\
}

\begin{document}

\maketitle

\begin{abstract}
 Inverse reinforcement learning has proved its ability to explain state-action trajectories of expert agents by recovering their underlying reward functions in increasingly challenging environments. Recent advances in adversarial learning have allowed extending inverse RL to applications with non-stationary environment dynamics unknown to the agents, arbitrary structures of reward functions and improved handling of the ambiguities inherent to the ill-posed nature of inverse RL. This is particularly relevant in real time applications on stochastic environments involving risk, like volatile financial markets. Moreover, recent work on simulation of complex environments enable learning algorithms to engage with real market data through simulations of its latent space representations, avoiding a costly exploration of the original environment. In this paper, we explore whether adversarial inverse RL algorithms can be adapted and trained within such latent space simulations from real market data, while maintaining their ability to recover agent rewards robust to variations in the underlying dynamics, and transfer them to new regimes of the original environment.
 
\end{abstract}

\section{Introduction}

Reinforcement learning (RL) achieves robust performance in a wide number of fields, with particularly relevant success in model-free applications \cite{mnih2013playing, vanHasselt2015dqn} where agents explore an environment with no prior knowledge about its underlying dynamics, and learn a policy that maximizes certain cumulative reward function. Such learning process typically requires recurrent access of the agent to the environment on a trial-and-error based exploration; however, reinforcement learning in risk-critical tasks such as automatic navigation or financial risk control would not allow such an exploration, since decisions have to be made in real time in a non-stationary environment where the risks and costs inherent to a trial-and-error approach can be unaffordable. In addition, such RL performance generally requires that the designer manually specifies a reward function that represents the task adequately and can be optimized efficiently \cite{ng1999policyinvariance}. 

In the context of learning from expert demonstrations, inverse reinforcement learning has proved capable of recovering through inference the reward function of expert agents through observations of their state-action trajectories \cite{ziebart2008maximum, levine2011nonlinear} with decreasing dependence on pre-defined assumptions about linearity or the general structure of the underlying reward function, generally under a maximum entropy framework \cite{ziebart2010modeling}. 

Recent advances in inverse RL have extended its application to high-dimensional state spaces with unknown dynamics and arbitrary non-linear reward structures, thanks to the use of neural networks to represent the reward function \cite{finn2016guided}. Adversarial inverse reinforcement learning (AIRL) \cite{fu2017learning} extends inverse RL further, achieving the recovery of rewards robust to variations in the dynamics of the environment, while learning at the same time a policy to perform the task; AIRL builds on the equivalences found by \cite{FinnCAL16} between inverse RL under maximum entropy and the training process of generative adversarial networks (GANs) \cite{goodfellow2014gans}. 

\subsection{Contributions} 
Financial markets are a particularly challenging case for inverse RL: high-dimensional, stochastic environments with non-stationary transition dynamics unknown to the observer, and with variable reactions to actions from agents. This makes AIRL particularly interesting to test on real financial data, aiming at learning from experts robust reward functions that can then be transferred to new regimes of the original environment. 

In this paper we explore the adaptation of AIRL to a volatile financial environment based on real tick data from a limit order book (LOB) in the stock market, attempting to recover the rewards from three expert market agents through an observer with no prior knowledge of the underlying dynamics, where such dynamics can also change with time following real market data, and where the environment reacts to the agent's actions. This is a challenging environment where we would expect AIRL to remain better suited to learn from the expert agent trajectories than methods aimed directly at recovering the policy (generative adversarial imitation learning \cite{ermon2016gail}, or GAIL), given the expected added value of AIRL regarding robustness of the recovered rewards with respect to varying dynamics of the environment. GAIL provides means analogous to generative adversarial networks that allow extraction of policies directly from data through a model-free approach for complex behaviours in high-dimensional environments. The performance of each method is then evaluated against the proportion of the expert agent's total cumulative reward that can be obtained by policies recovered through adversarial learning.

\subsection{Model environment based on real data}

The adversarial learning algorithms used in the experiment will require a model of the environment where the observed agent trajectories took place, in order to evaluate the iterative estimations of rewards and policies most likely to have generated the observations. In practice, we would observe expert trajectories from agents as training data for adversarial learning, and then transfer the learnt policies to new test market data from the real environment. 

The latent space model of the equity LOB described in  \cite{yuanbo2019} shows how the Mixture-Density Recurrent Network (RNN-MDN) architecture presented in World Models \cite{ha2018worldmodels} to learn the transition dynamics of the environment can also be applied to financial time series, learning a causal representation of LOB market data where RL experts can be trained to learn policies then transferable back to the original environment. Building on this work, we train three expert traders in the latent space market model through advantage actor critic (A2C) \cite{mnih2019a3c}, double DQN \cite{vanHasselt2015dqn}, and Policy Gradient \cite{williams92policygradient} respectively, whose learnt policies remain profitable when tested on subsequent time series out of sample. 
The collections of expert state-action trajectories generated by each trained agent serve as input for the AIRL and GAIL algorithms in our experiments, following the implementation by \cite{fu2017learning}. Our conclusions will then examine the proportion of the experts' cumulative rewards produced by the policies learnt through either AIRL or GAIL from each expert agent.

\section{Background} \label{sec:background}

\subsection{Related work}

A number of previous works have applied inverse RL to financial data, focusing on evaluations of feature vectors for state representations at different scales to explore a market of competing agents \cite{hendricks2017}, and assuming linear structures for the reward functions. Other works have focused on assessing the comparative performance of probabilistic inverse RL based on Bayesian neural networks \cite{roavicens2019} as an alternative to inverse RL based on Gaussian processes \cite{levine2011nonlinear}, applied to a simulated financial market under a maximum entropy framework and allowing non-linear structures in the reward functions of the agents.

On the other hand, model-based approaches have attempted to recover specific parameters such as risk aversion implied by data \cite{halperin2018}, where the observer assumes a certain structure for the underlying utility of the agent. Research with simulations of real environments through neural networks \cite{kaiser2019mbrl} allows to extend the original action and reward spaces to produce observations in the same spaces. The representation of an environment through generative models has also been previously described by World Models \cite{ha2018worldmodels} and its adaptation to limit order books \cite{yuanbo2019}, where the authors obtain latent representations of the environment enabling agents to learn a policy efficiently, and to transfer it back to the original environment. Other authors have explored applications of Gaussian inverse RL to learning investor sentiment models from historic data \cite{yangYA18}, to then extract tradable signals from the model.

\subsection{Adversarial Learning under Maximum Causal Entropy} \label{sub:irl}

The adversarial IRL experiments presented in this paper follow the AIRL implementation described in \cite{fu2017learning}, framed through a standard Markov decision process defined by a tuple $\langle \mathcal{S},\mathcal{A},\mathcal{T},r,\gamma \rangle$ consisting of a state space  $\mathcal{S}$; a set  $\mathcal{A}$ of eligible actions; a model $\mathcal{T}$ of transition dynamics $\mathcal{T} (s', a, s)$, where each $p(s'|s,a)$ represents the transition probability from state $s$ through action $a$ to state $s'$; the unknown reward function $r(s, a)$ that we aim at recovering; and the discount factor $\gamma$ taking values between [0, 1]. 

In general terms, forward RL seeks an optimal policy $\pi^{*}$  that maximizes an expected cumulative reward $\mathbb{E}\big[ \sum_{t=0}^{T} \gamma^t r(s_t)|\pi^*\big]$ under the transition dynamics reflected in $\mathcal{T} (s', a, s)$ and the policy $\pi^*$. We denote the state-action trajectories derived from such expert policy as $\mathbf{x} = \{(s_t, a_t)\}_{t=0}^{T}$. 

Under this framework, inverse RL attempts to recover the reward function $r$ that would most likely have generated a given set of expert trajectories $\mathcal{D}=\{\mathbf{x}_{n}\}_{n=1}^{N}$ under the MDP $\langle \mathcal{S},\mathcal{A},\mathcal{T},\gamma \rangle$ of unknown reward. The adversarial learning methods considered will observe the trajectories in $\mathcal{D}$ to infer a reward $\hat{r}$ (that yields a policy $\hat{\pi}$) to explain $\mathcal{D}$. The performance of each model is then evaluated through the total cumulative reward $\mathbb{E}\big[ \sum_{t=0}^{T} \gamma^t r(s_t)|\hat{\pi}\big]$ that $\hat{\pi}$ can recover against the total reward $\mathbb{E}\big[ \sum_{t=0}^{T} \gamma^t r(s_t)|\pi^*\big]$ obtained by the expert agent under policy $\pi^*$.

One of the main challenges of inverse RL resides in its ill-posed nature: firstly, a given set of state-action trajectories $\mathcal{D}$ may  be explained by several different optimal policies \cite{ng1999policyinvariance}; the maximum entropy principle as proposed in \cite{ziebart2008maximum} addresses this problem, assuming that the optimal policy $\pi^{*}(a | s)$ that would have generated the trajectories in $\mathcal{D}$ follows $\exp \left\{Q'\left(s, a\right)\right\}$ as developed in \cite{ziebart2008maximum, haarnoja2017reinforcement}. The soft $Q'$ function refers to the learning process under the standard RL formulation, where the objective is regularized against a metric of differential entropy.

Secondly, the optimal policy determined may also be explained by a number of different reward functions, where an external observer lacks the means to distinguish true rewards from those product of non-stationary environment dynamics. The latter, when transferred to an environment with new dynamics, may fail to produce an optimal outcome. Adversarial inverse RL \cite{fu2017learning} introduces the concept of \textit{disentangled rewards}, aiming at learning reward functions robust to variations in the environmental dynamics. This makes AIRL particularly attractive to study agents in financial markets, given their inherent need of continuous adaptation to changing dynamics.
 
The connection between inverse RL under maximum causal entropy and GANs as described by \cite{FinnCAL16} compares the iterative cycles between generator and discriminator in the GAN with cases of inverse RL that employ neural nets to learn generic reward functions under unknown environment dynamics \cite{finn2016guided, boularias2011a}. Moreover, a probabilistic approach to the problem allows to capture the stochastic nature of the relationship between the reward objective and the policy actually executed. 

Following the cited previous works, we use as benchmark a GAIL implementation adapted to our market model: while GAIL is able to learn a policy from expert demonstrations, we expect AIRL to outperform in an environment with volatile, non-stationary dynamics, besides recovering a reward function in addition to the policy.

\section{Experimental setup and results}

The first requirement of our experiments is a model environment based on real financial data, that allows training of RL agents and is also compatible with the AIRL and GAIL learning algorithms. Learning a rich representation of the environment  adds the general advantage of allowing RL models that are simpler, smaller and less expensive to train than model-free counterparts for a certain target performance of the learnt policy, as they search in a smaller space. Our experimental setup is based on an OpenAI Gym structure \cite{brockman2016openai} used to adapt the AIRL and GAIL implementations by \cite{fu2017learning} to a financial version of a World Model environment as follows:

\subsection{World models learnt from observed environments\label{subsection:WM}}

World Models \cite{ha2018worldmodels} provides an unsupervised probabilistic generative method to learn world representations in space and time through neural networks, including the transition dynamics as our experiment requires. RL agents operating in the real world would adapt their policies to their expectations of such transition dynamics, i.e. probabilities $p(s'|s,a)$ of certain future state $s'$ given the current state $s$ and choice of action $a$ contained in $\mathcal{T} (s', a, s)$. RL agents can then be trained within the world model, with their learnt policies remaining optimal once transferred back to the original environment (or even outperforming the agents trained in the original environment in some instances). 

\begin{figure}[!htb] 
    \centering
    \includegraphics[width=1.0\textwidth]{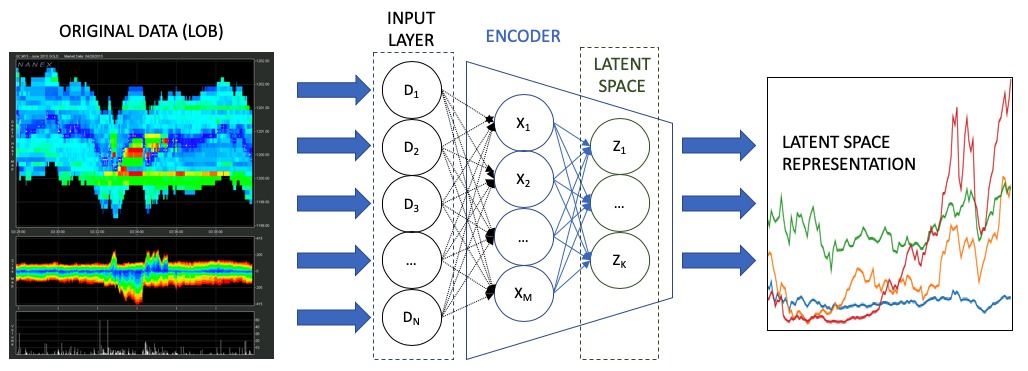} 
    \caption{An auto-encoder extracts a latent-space representation from the original LOB data \cite{nxcore_nanex} \label{fig:ave}}
\end{figure}

These models capture three elements of the original environment: firstly, an 
auto-encoder learns a latent representation of each state vector from its original high-dimensional space, compressing all the observed information of that state into a vector $z$ within a latent space of lower dimension (Fig. \ref{fig:ave}).

Secondly, after learning the latent vector for each state, the model learns the transition dynamics  connecting consecutive states through certain action choices $a_t$. This model of transition probabilities $p(s'|s,a)$ of $\mathcal{T} (s', a, s)$ in the MDP is learnt through an RNN \cite{Schmidhuber90makingthe, Schmidhuber90anon, Schmidhuber91nips990_393}, able to learn sequential information from time series along various time horizons by learning $P(z_{t+1}|a_t, h_t, z_t)$ where $h_t$ represents a hidden state of the RNN. 

The choice of RNN architectures is based on their suitability to capture the transition model of sequential data \cite{Ha_Eck2017}, thus learning some predictive modelling of future states. In order to better reflect the stochastic nature of the target environment, the deterministic output of the RNN is then fed into a Mixture Density Network (MDN) \cite{Bishop94mixturedensity} to produce a probabilistic prediction of next latent state. This combined RNN-MDN architecture is detailed further in \cite{ha2018worldmodels, graves2013}. 

Finally, an RL module decides which action to take given all the above information, including the transition predicted in the latent space. As noted in \cite{ha2018worldmodels}, this approach has the additional advantage of providing RL agents with access to the latent space learnt by the model, which includes representations of both the current state and an indication of what to expect in upcoming states through the transitions modelled, sampling from the probability distribution of future states provided by the RNN-MDN.

\begin{figure}[!htb]    
    \centering
    \includegraphics[width=1.0\textwidth]{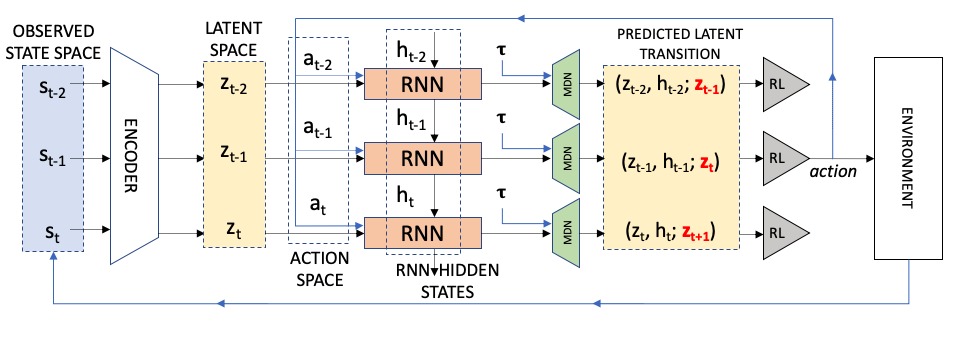}
    \caption{Model for prediction of latent space transitions. Our RL processes are run in the latent state space. \label{fig:rnn}}
\end{figure}

\subsection{LOB version of world models learnt from real market data\label{subsection:finWM}}

The original version of World Models learns from environments based on computer simulations, such as {\fontfamily{qcr}\selectfont
CarRacing-v0} \cite{klimov2016} and VizDoom \cite{kempkaWRTJ16vizdoom}. We base our experiments on the extension of this architecture to financial environments presented in \cite{yuanbo2019}, where the auto-encoder is trained to learn a latent space from sequences of snapshots of actual limit order book data, and the RNN-MDN models the transitions between consecutive time frames as follows:

The data used to train the model consists of LOB time series from shares of Tencent Holdings Ltd. (HK-700) traded in the Hong Kong stock exchange along sixty trading days between January and March, 2018. Data from the following twenty trading days in April is then used as testing reference for the adversarial learning algorithms. Each state contains the sequence of the last 10 data ticks for 3 LOB levels, so that the sequential information necessary to learn the transition dynamics is captured in the data for each state.

LOB data is of high dimension, generally denominated Level II market data, and features several relevant data points for each timestamp. Firstly, it contains various levels of bid (buy) and ask (sell) prices for each timestamp, typically ranked in decreasing order of competitivity. The \textit{mid price} is generally derived as an average between the highest bid and lowest ask in the first LOB level. Secondly, it includes the trading volume associated with each of such prices offered. The combined progression in time of this data structure is often represented as a tensor of three dimensions \cite{tran2017hfttensors}. Finally, trade stamp series contain the price and size of the last transactions executed out of previous LOB states, used in this model as RL exploration.

In  order to allow the RNN-MDN to model the transition dynamics, each state $s_t$ consists of 10 consecutive LOB data ticks for each of the above features. The auto-encoder learns a compressed version of the data in lower dimensionality, through a latent space representation of short sequences of original market data (Fig. \ref{fig:ave}). The model used in \cite{yuanbo2019} features a CNN-based auto-encoder, looking back 10 updates in the limit order book that contains 3 levels of orders, and condenses the $(10*3*4)$ matrix into a $z$ vector of dimension 12. 

The RNN-MDN used to model the transition dynamics assumes that $p(s'| s,a)$ follows a Gaussian mixture distribution \cite{yuanbo2019,ha2018worldmodels}, where the model learns the parameters of the distribution $p(s'| s,a) = \sum_{k=t}^{K} w_k (s,a) \mathcal{D}(s'|\mu_k (s,a), \sigma_k ^2 (s,a))$ with $\mathcal{D(\cdot)}$ assumed to be Gaussian, $k=5$ and 128 neurons per layer in the RNN. Once the prediction of the next state is sampled in the latent space, the reward with respect to the present state and the action chosen is produced through a regression model.

\subsection{Training experts in the LOB latent space} 

Once the full world model is trained and integrated into an OpenAI Gym structure, we train in it the expert agents whose trajectories will serve as input to the adversarial learning experiments. We follow the selection of RL algorithms in \cite{yuanbo2019} proved to learn policies in the world model that remain profitable when transferred back to the real environment: double DQN \cite{vanHasselt2015dqn}, advantage actor critic (A2C) \cite{mnih2019a3c} and policy gradient \cite{williams92policygradient}. 

We now use each of the three trained expert agents to generate collections of 100 state-action trajectories for each agent, with a length of 1000 $(s_t, a_t)$ pairs each. Among the three experts, A2C seems to be the best performing agent in terms of high mean and low variance of cumulative rewards, followed by DQN (Fig. \ref{fig:expert_rewards}):

\begin{figure}[!htb]    
    \centering
    \includegraphics[width=0.8\textwidth]{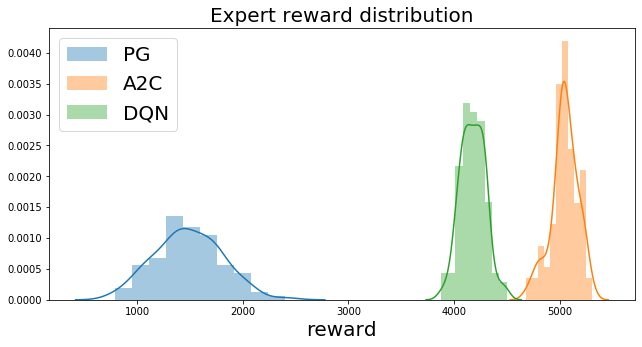} 
    \caption{Distribution of rewards obtained by each expert RL agent. \label{fig:expert_rewards} }
\end{figure}

\subsection{Adversarial learning from Reinforcement Learning trading experts}

As in general GAN structures, adversarial inverse reinforcement learning is implemented through a generator and a discriminator: two neural networks that contest with each other. The generator learns to produce an output mapping from a latent space to a target data distribution, while the discriminator tries to classify which samples come from the true distribution and which where produced by the generator.

In the case of AIRL, the generator learns to produce state-action pairs ${(s_t, a_t)}^L_i$ based on those observed in the trajectories from the expert demonstrations ${(s_t, a_t)}^E_i$, while the discriminator $D_\theta$ tries to classify which state-action pairs were produced by the generator against those actually produced by an expert agent, training through binary logistic regression. The reward estimate $\hat{r}_\theta$ is then updated from the GAN minimax loss: $\log D_\theta (s,a,s') - \log(1-D_\theta(s,a,s'))$. An estimate $\hat{\pi}$ of the policy is then updated from $\hat{r}_\theta$ iteratively.

Our experiments are initialized with the expert trajectories ${(s_t, a_t)}^E_i$ produced by executing each expert policy $\pi^E$ within the world model learnt from data gathered between January and March. We then initialize a policy $\hat{\pi}_{0}$ and the discriminator $D_\theta$. The adversarial observer then updates iteratively $D_\theta$ and the reward $\hat{r}_\theta$ that produces a policy $\hat{\pi}$, which generates samples of ${(s_t, a_t)}^L_i$ increasingly similar to ${(s_t, a_t)}^E_i$. The evaluations of cumulative rewards produced by each $\hat{\pi}$ take place with a model environment run with test data from April. 

\begin{figure*}[h!]
\centering
    \begin{subfigure}[b]{0.5\linewidth}
      \centering
      \includegraphics[width=\linewidth]{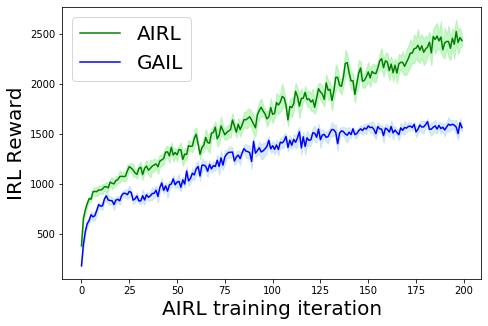}
      \caption{Trained with A2C Experts}
      \label{irlreturn_a2c}
    \end{subfigure}%
    \begin{subfigure}[b]{0.5\linewidth}
      \centering
      \includegraphics[width=\linewidth]{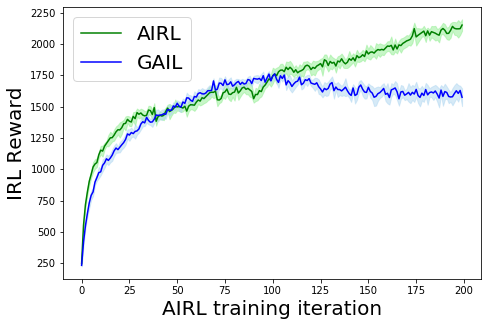}
      \caption{Trained with DQN Experts}
      \label{irlreturn_dqn}
    \end{subfigure}
    \begin{subfigure}[b]{0.5\linewidth}
      \centering
      \includegraphics[width=\linewidth]{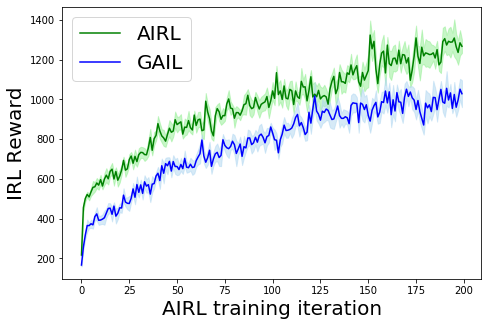}
      \caption{Trained with PG Experts}
      \label{irlreturn_pg}
    \end{subfigure}

    \caption{Comparative of rewards obtained through AIRL and GAIL by learning from each expert agent.}
\label{irl_reward}
\end{figure*}

We present in Fig. \ref{irl_reward} and Table \ref{agent_performance} the results of running AIRL and GAIL on 200 training iterations: AIRL is found to outperform GAIL in all the three cases considered (learning from demonstrations of each expert trained), consistently with the initial motivation of the experiment.

\begin{table}[h!]
\centering
    \begin{subtable}{.8\textwidth}
        \centering
        \begin{tabular}{|l|l|l|}
        \hline
        \centering
        Agent     & Average Return & Std Return  \\ \hline
        Expert A2C  & 5047.3      & 125.9    \\ \hline
        AIRL        & 2433.0      & 59.7    \\ \hline
        GAIL        & 1563.1       & 43.4    \\ \hline
        \end{tabular}
        \caption{Adversarial learning from A2C Experts}
    \end{subtable}%
    \hfill
    \begin{subtable}{.8\textwidth}
        \centering
        \begin{tabular}{|l|l|l|}
        \hline
        \centering
        Agent     & Average Return & Std Return  \\ \hline
        Expert DQN  & 4177.4      & 116.6    \\ \hline
        AIRL        & 2151.1      & 45.6    \\ \hline
        GAIL        & 1574.5      & 70.8     \\ \hline
        \end{tabular}
        \caption{Adversarial learning from DQN Experts}
    \end{subtable}%
    \hfill
    \begin{subtable}{.8\textwidth}
        \centering
        \begin{tabular}{|l|l|l|}
        \hline
        \centering
        Agent     & Average Return & Std Return  \\ \hline
        Expert PG  & 1483.3      & 313.2    \\ \hline
        AIRL        & 1267.7      & 33.8    \\ \hline
        GAIL        & 1029.1      & 67.9     \\ \hline
        \end{tabular}
        \caption{Adversarial learning from PG Experts}
    \end{subtable}%
    \hfill

\caption{Summary of cumulative rewards obtained by AIRL, GAIL and each expert agent in the order book.}
\label{agent_performance}
\end{table}

\newpage
\clearpage
\section{Conclusions}

We have presented an experimental setup to adapt adversarial inverse reinforcement learning and generative adversarial imitation learning to a volatile financial environment, in order to evaluate their ability to learn from expert trajectories in latent space simulations: the latent model employed allows the training of the adversarial algorithms based on real market data. The results obtained show that both methods can be trained against a latent space model of the market, while the specific advantage of AIRL for learning rewards robust to variable environment dynamics allows it to outperform GAIL in the cumulative rewards recovered from each of the three agents as the number of iterations increases. 

\begin{figure*}[h!]
\centering
    \begin{subfigure}[b]{0.5\linewidth}
      \centering
      \includegraphics[width=\linewidth]{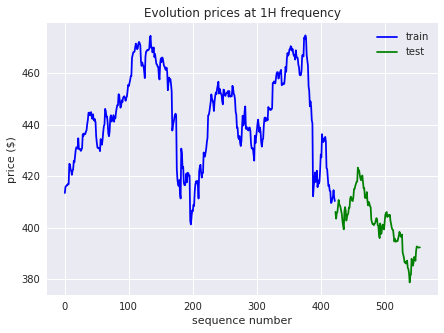}
      \caption{Mid price evolution, sampled hourly.}
      \label{irlreturn_a2c}
    \end{subfigure}%
    \begin{subfigure}[b]{0.5\linewidth}
      \centering
      \includegraphics[width=\linewidth]{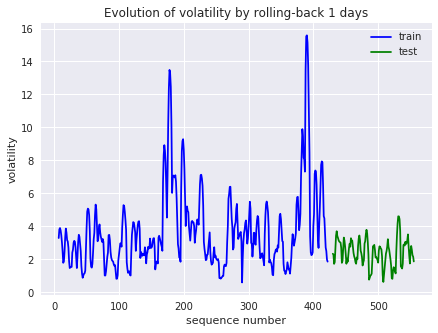}
      \caption{Volatility of mid prices (1-day window)}
      \label{irlreturn_dqn}
    \end{subfigure}
    \begin{subfigure}[b]{0.5\linewidth}
      \centering
      \includegraphics[width=\linewidth]{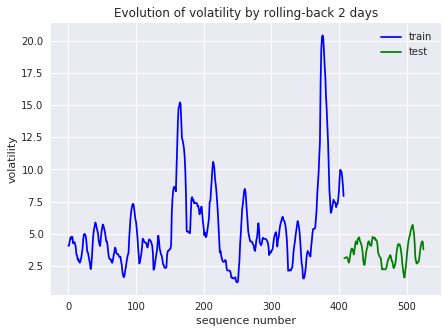}
      \caption{Volatility of mid prices (2-day window)}
      \label{irlreturn_pg}
    \end{subfigure}

    \caption{The original data series used to train the latent space model and the experts (blue) and to test the adversarial learning methods (green) contain significant variability of price and volatility levels.}
\label{env_variability}
\end{figure*}

A review of the original data series (Fig. \ref{env_variability}) confirms the variability of mid price series and of their volatility levels (measured across various horizons) when comparing the data corresponding to the training period against that of the testing period. The rewards learnt by AIRL from the training model are robust to such variable dynamics, hence its outperformance over GAIL when evaluated on the test period.

\newpage
\clearpage

\bibliography{neurips_2019}

\begin{thebibliography}{10}

\bibitem{mnih2013playing}
Volodymyr Mnih, Koray Kavukcuoglu, David Silver, Alex Graves, Ioannis
  Antonoglou, Daan Wierstra, and Martin Riedmiller.
\newblock Playing atari with deep reinforcement learning.
\newblock {\em arXiv preprint arXiv:1312.5602}, 2013.

\bibitem{vanHasselt2015dqn}
Hado van Hasselt, Arthur Guez, and David Silver.
\newblock Deep reinforcement learning with double q-learning.
\newblock {\em CoRR}, abs/1509.06461, 2015.

\bibitem{ng1999policyinvariance}
Andrew~Y. Ng, Daishi Harada, and Stuart~J. Russell.
\newblock Policy invariance under reward transformations: Theory and
  application to reward shaping.
\newblock In {\em Proceedings of the Sixteenth International Conference on
  Machine Learning}, ICML '99, pages 278--287, 1999.

\bibitem{ziebart2008maximum}
Brian~D Ziebart, Andrew~L Maas, J~Andrew Bagnell, and Anind~K Dey.
\newblock Maximum entropy inverse reinforcement learning.
\newblock In {\em AAAI}, volume~8, pages 1433--1438. Chicago, IL, USA, 2008.

\bibitem{levine2011nonlinear}
Sergey Levine, Zoran Popovic, and Vladlen Koltun.
\newblock Nonlinear inverse reinforcement learning with gaussian processes.
\newblock In {\em Advances in Neural Information Processing Systems}, pages
  19--27, 2011.

\bibitem{ziebart2010modeling}
Brian~D. Ziebart.
\newblock {\em Modeling Purposeful Adaptive Behavior with the Principle of
  Maximum Causal Entropy}.
\newblock PhD thesis, Pittsburgh, PA, USA, 2010.
\newblock AAI3438449.

\bibitem{finn2016guided}
Chelsea Finn, Sergey Levine, and Pieter Abbeel.
\newblock Guided cost learning: Deep inverse optimal control via policy
  optimization.
\newblock In {\em International Conference on Machine Learning}, pages 49--58,
  2016.

\bibitem{fu2017learning}
Justin Fu, Katie Luo, and Sergey Levine.
\newblock Learning robust rewards with adversarial inverse reinforcement
  learning.
\newblock {\em arXiv preprint arXiv:1710.11248}, 2017.

\bibitem{FinnCAL16}
Chelsea Finn, Paul~F. Christiano, Pieter Abbeel, and Sergey Levine.
\newblock A connection between generative adversarial networks, inverse
  reinforcement learning, and energy-based models.
\newblock {\em CoRR}, abs/1611.03852, 2016.

\bibitem{goodfellow2014gans}
Ian Goodfellow, Jean Pouget-Abadie, Mehdi Mirza, Bing Xu, David Warde-Farley,
  Sherjil Ozair, Aaron Courville, and Yoshua Bengio.
\newblock Generative adversarial nets.
\newblock In Z.~Ghahramani, M.~Welling, C.~Cortes, N.~D. Lawrence, and K.~Q.
  Weinberger, editors, {\em Advances in Neural Information Processing Systems
  27}, pages 2672--2680. Curran Associates, Inc., 2014.

\bibitem{ermon2016gail}
Jonathan Ho and Stefano Ermon.
\newblock Generative adversarial imitation learning.
\newblock In D.~D. Lee, M.~Sugiyama, U.~V. Luxburg, I.~Guyon, and R.~Garnett,
  editors, {\em Advances in Neural Information Processing Systems 29}, pages
  4565--4573. Curran Associates, Inc., 2016.

\bibitem{yuanbo2019}
Haoran Wei, Yuanbo Wang, Lidia Mangu, and Keith Decker.
\newblock Model-based reinforcement learning for predictions and control for
  limit order books, 2019.
\newblock Cite Arxiv 1910.03743.

\bibitem{ha2018worldmodels}
David Ha and J{\"{u}}rgen Schmidhuber.
\newblock World models.
\newblock {\em CoRR}, abs/1803.10122, 2018.

\bibitem{mnih2019a3c}
Volodymyr Mnih, Adria~Puigdomenech Badia, Mehdi Mirza, Alex Graves, Timothy
  Lillicrap, Tim Harley, David Silver, and Koray Kavukcuoglu.
\newblock Asynchronous methods for deep reinforcement learning.
\newblock In Maria~Florina Balcan and Kilian~Q. Weinberger, editors, {\em
  Proceedings of The 33rd International Conference on Machine Learning},
  volume~48 of {\em Proceedings of Machine Learning Research}, pages
  1928--1937, New York, New York, USA, 20--22 Jun 2016. PMLR.

\bibitem{williams92policygradient}
Ronald~J. Williams.
\newblock Simple statistical gradient-following algorithms for connectionist
  reinforcement learning.
\newblock In {\em Machine Learning}, pages 229--256, 1992.

\bibitem{hendricks2017}
Dieter Hendricks, Adam Cobb, Richard Everett, Jonathan Downing, and Stephen~J.
  Roberts.
\newblock {Inferring agent objectives at different scales of a complex adaptive
  system}.
\newblock Papers 1712.01137, arXiv.org, December 2017.

\bibitem{roavicens2019}
Jacobo Roa~Vicens, Cyrine Chtourou, Angelos Filos, Francisco Rullan, Yarin Gal,
  and Ricardo Silva.
\newblock Towards inverse reinforcement learning for limit order book dynamics.
\newblock In {\em ICML Workshop 'AI in Finance: Applications and Infrastructure
  for Multi-Agent Learning' at the 36th International Conference on Machine
  Learning}, 2019.

\bibitem{halperin2018}
Igor Halperin and Ilya Feldshteyn.
\newblock Market self-learning of signals, impact and optimal trading:
  Invisible hand inference with free energy.
\newblock {\em arXiv preprint arXiv:1805.06126}, 2018.

\bibitem{kaiser2019mbrl}
Lukasz Kaiser, Mohammad Babaeizadeh, Piotr Milos, Blazej Osinski, Roy~H.
  Campbell, Konrad Czechowski, Dumitru Erhan, Chelsea Finn, Piotr Kozakowski,
  Sergey Levine, Ryan Sepassi, George Tucker, and Henryk Michalewski.
\newblock Model-based reinforcement learning for atari.
\newblock {\em CoRR}, abs/1903.00374, 2019.

\bibitem{yangYA18}
Steve~Y. Yang, Yangyang Yu, and Saud Almahdi.
\newblock An investor sentiment reward-based trading system using gaussian
  inverse reinforcement learning algorithm.
\newblock {\em Expert Syst. Appl.}, 114:388--401, 2018.

\bibitem{haarnoja2017reinforcement}
Tuomas Haarnoja, Haoran Tang, Pieter Abbeel, and Sergey Levine.
\newblock Reinforcement learning with deep energy-based policies.
\newblock In {\em Proceedings of the 34th International Conference on Machine
  Learning-Volume 70}, pages 1352--1361. JMLR. org, 2017.

\bibitem{boularias2011a}
Abdeslam Boularias, Jens Kober, and Jan Peters.
\newblock Relative entropy inverse reinforcement learning.
\newblock In Geoffrey Gordon, David Dunson, and Miroslav Dudík, editors, {\em
  Proceedings of the Fourteenth International Conference on Artificial
  Intelligence and Statistics}, volume~15 of {\em Proceedings of Machine
  Learning Research}, pages 182--189, Fort Lauderdale, FL, USA, 11--13 Apr
  2011. PMLR.

\bibitem{brockman2016openai}
Greg Brockman, Vicki Cheung, Ludwig Pettersson, Jonas Schneider, John Schulman,
  Jie Tang, and Wojciech Zaremba.
\newblock Openai gym, 2016.
\newblock Cite arxiv:1606.01540.

\bibitem{nxcore_nanex}
LOB representation~from NxCore~API.
\newblock 2019.
\newblock Nanex Corp.

\bibitem{Schmidhuber90makingthe}
Jürgen Schmidhuber.
\newblock Making the world differentiable: On using self-supervised fully
  recurrent neural networks for dynamic reinforcement learning and planning in
  non-stationary environments.
\newblock Technical report, 1990.

\bibitem{Schmidhuber90anon}
Jürgen Schmidhuber.
\newblock An on-line algorithm for dynamic reinforcement learning and planning
  in reactive environments.
\newblock In {\em In Proc. IEEE/INNS International Joint Conference on Neural
  Networks}, pages 253--258. IEEE Press, 1990.

\bibitem{Schmidhuber91nips990_393}
J\"{u}rgen Schmidhuber.
\newblock Reinforcement learning in markovian and non-markovian environments.
\newblock In R.~P. Lippmann, J.~E. Moody, and D.~S. Touretzky, editors, {\em
  Advances in Neural Information Processing Systems 3}, pages 500--506.
  Morgan-Kaufmann, 1991.

\bibitem{Ha_Eck2017}
David Ha and Douglas Eck.
\newblock A neural representation of sketch drawings.
\newblock {\em CoRR}, abs/1704.03477, 2017.

\bibitem{Bishop94mixturedensity}
Christopher~M. Bishop.
\newblock Mixture density networks.
\newblock Technical report, 1994.

\bibitem{graves2013}
Alex Graves.
\newblock Generating sequences with recurrent neural networks.
\newblock {\em CoRR}, abs/1308.0850, 2013.

\bibitem{klimov2016}
Oleg Klimov.
\newblock Carracing-v0.
\newblock 2016.
\newblock \url{https://gym.openai.com/envs/CarRacing-v0/}.

\bibitem{kempkaWRTJ16vizdoom}
Michal Kempka, Marek Wydmuch, Grzegorz Runc, Jakub Toczek, and Wojciech
  Jaskowski.
\newblock Vizdoom: {A} doom-based {AI} research platform for visual
  reinforcement learning.
\newblock {\em CoRR}, abs/1605.02097, 2016.

\bibitem{tran2017hfttensors}
Dat~Thanh Tran, Martin Magris, Juho Kanniainen, Moncef Gabbouj, and Alexandros
  Iosifidis.
\newblock Tensor representation in high-frequency financial data for price
  change prediction.
\newblock {\em CoRR}, abs/1709.01268, 2017.

\end{thebibliography}
\bibliographystyle{unsrt}

\textbf{Disclaimer}

Opinions and estimates constitute our judgement as of the date of this Material, are for informational purposes only and are subject to change without notice. This Material is not the product of J.P. Morgan’s Research Department and therefore, has not been prepared in accordance with legal requirements to promote the independence of research, including but not limited to, the prohibition on the dealing ahead of the dissemination of investment research. This Material is not intended as research, a recommendation, advice, offer or solicitation for the purchase or sale of any financial product or service, or to be used in any way for evaluating the merits of participating in any transaction. It is not a research report and is not intended as such. Past performance is not indicative of future results. Please consult your own advisors regarding legal, tax, accounting or any other aspects including suitability implications for your particular circumstances. J.P. Morgan disclaims any responsibility or liability whatsoever for the quality, accuracy or completeness of the information herein, and for any reliance on, or use of this material in any way.

Important disclosures at: www.jpmorgan.com/disclosures

\end{document}